\documentstyle[aclap,psfig]{article}

\def\W{{\cal W}}
\def\given{\,|\,}

\def\pt{\tilde p}

\def\p{{p}}

\def\A{A}

\def\W{{W}}

\def\pt{\tilde p}
\def\LL{{\cal L}}
\def\pmean#1#2{\langle\, #2\, \rangle_{#1}}
\def\pmean#1#2{E_{#1}\left[ #2 \right]}
\def\given{\,|\,}
\def\pts{p_{\mu_1,\mu_2}}
\def\ptsp{p_{\mu_1^\prime,\mu_2^\prime}}
\def\A{{\cal A}}
\def\W{N}

\def\p{p}

\title{A Model of Lexical Attraction and Repulsion\thanks{Research
supported in part by NSF grant IRI-9314969,
DARPA AASERT award DAAH04-95-1-0475, and the 
ATR Interpreting Telecommunications Research Laboratories.}}
\author{{\bf Doug Beeferman \quad Adam Berger \quad John Lafferty} \\[5pt]
School of Computer Science \\ 
Carnegie Mellon University \\
Pittsburgh, PA 15213  USA \\[5pt]
{\tt <dougb,aberger,lafferty>@cs.cmu.edu}}

\begin{document}
\bibliographystyle{acl}
\maketitle
\vspace{-0.5in}

\begin{abstract}
This paper introduces new methods based on exponential families for
modeling the correlations between words in text and speech.  While
previous work assumed the effects of word co-occurrence statistics to
be constant over a window of several hundred words, we show that their
influence is nonstationary on a much smaller time scale.  Empirical
data drawn from English and Japanese text, as well as conversational
speech, reveals that the ``attraction'' between words decays
exponentially, while stylistic and syntactic contraints create a
``repulsion'' between words that discourages close co-occurrence.  We
show that these characteristics are well described by simple mixture
models based on two-stage exponential distributions which can be
trained using the EM algorithm.  The resulting distance distributions
can then be incorporated as penalizing features in an exponential
language model.
\end{abstract}

\section{Introduction}

One of the fundamental characteristics of language, viewed as a stochastic
process, is that it is highly {\it nonstationary\/}.  Throughout a written
document and during the course of spoken conversation, the topic evolves,
effecting local statistics on word occurrences. The standard trigram
model disregards this nonstationarity, as does any stochastic
grammar which assigns probabilities to sentences in a context-independent
fashion. 

Stationary models are used to describe such a dynamic
source for at least two reasons. The first is convenience: stationary
models require a relatively small amount of computation to train and
to apply.  The second is ignorance: we know so little about how to
model effectively the nonstationary characteristics of language that
we have for the most part completely neglected the problem. From a
theoretical standpoint, we appeal to the Shannon-McMillan-Breiman
theorem \cite{Cover:91a} whenever computing perplexities on test data;
yet this result only rigorously applies to stationary and ergodic
sources.

To allow a language model to adapt to its recent context, some
researchers have used techniques to update trigram statistics in a
dynamic fashion by creating a cache of the most recently seen
$n$-grams which is smoothed together (typically by linear
interpolation) with the static model; see for example
\cite{Jelinek:91a,Kuhn:90}.  
Another approach, using maximum entropy methods similar to those that we
present here, introduces a parameter for {\it trigger pairs} of mutually
informative words, so that the occurrence of certain words in recent
context boosts the probability of the words that they trigger
\cite{Rosenfeld:96}.  Triggers have also been incorporated through different
methods \cite{Kuhn:90,Ney:94}.  
All of these techniques treat the recent
context as a ``bag of words,'' so that a word that appears, say, five
positions back makes the same contribution to prediction as words
at distances of $50$ or $500$ positions back in the history.

In this paper we introduce new modeling techniques based on exponential
families for capturing the long-range correlations between occurrences
of words in text and speech.  We show how for both written text and
conversational speech, the empirical distribution of the distance
between trigger words exhibits a striking behavior in which the
``attraction'' between words decays exponentially, while stylistic and
syntactic constraints create a 
``repulsion'' between words
that discourages close co-occurrence.  

We have discovered that this observed behavior is
well described by simple mixture models based on two-stage
exponential distributions. Though in common use in 
queueing theory, such distributions have not, to our knowledge,
been previously exploited in speech and language processing.  
It is remarkable that the behavior of a highly complex stochastic
process such as the separation between word co-occurrences is well
modeled by such a simple parametric family, just as it is surprising
that Zipf's law can so simply capture the distribution of word
frequencies in most languages.  


In the following section we present examples of the empirical evidence
for the effects of distance.  In Section~3 we outline the class of
statistical models that we propose to model this data.
After completing this work we learned of a related paper
\cite{Niesler:97} which constructs similar models.
In Section~4 we present a parameter estimation algorithm, based on 
the EM algorithm,
for determining the maximum likelihood estimates within the class.  
In Section~5 we explain how 
distance models can be incorporated into an exponential language model,
and present sample perplexity results we have obtained using this class
of models.

\section{The Empirical Evidence}

The work described in this paper began with the goal of building a
statistical language model using a static trigram model as a ``prior,''
or default distribution, and adding certain features to a family of
conditional exponential models to capture some of the nonstationary
features of text.  The features we used were simple ``trigger pairs''
of words that were chosen on the basis of mutual information.
Figure~\ref{tab:trigger-examples} provides a small sample of the $41,263$
$(s,t)$ trigger pairs used in most of the experiments we will describe.

\begin{figure}[ht]
\begin{center}
\tt
\begin{tabular}{|| l | l ||}
\hline
$s$ & $t$ \\
\hline
Ms.  & her                \\
changes & revisions       \\
energy & gas              \\
committee & representative\\
board & board             \\
lieutenant & colonel      \\
AIDS & AIDS               \\
Soviet & missiles         \\
underwater & diving       \\
patients & drugs          \\
television & airwaves     \\
Voyager & Neptune \\                  
medical & surgical        \\
I & me                    \\
Gulf & Gulf               \\
\hline
\end{tabular} 
\end{center}
\caption{A sample of the 41,263 trigger pairs extracted from the
38~million word Wall Street Journal corpus.}
\label{tab:trigger-examples}
\end{figure}

\def\entry#1#2#3#4{#1 & #3 \\[-4pt] \small\tt #2 & \small\tt #4\\}

\begin{figure}[ht]
\centerline{\psfig{file=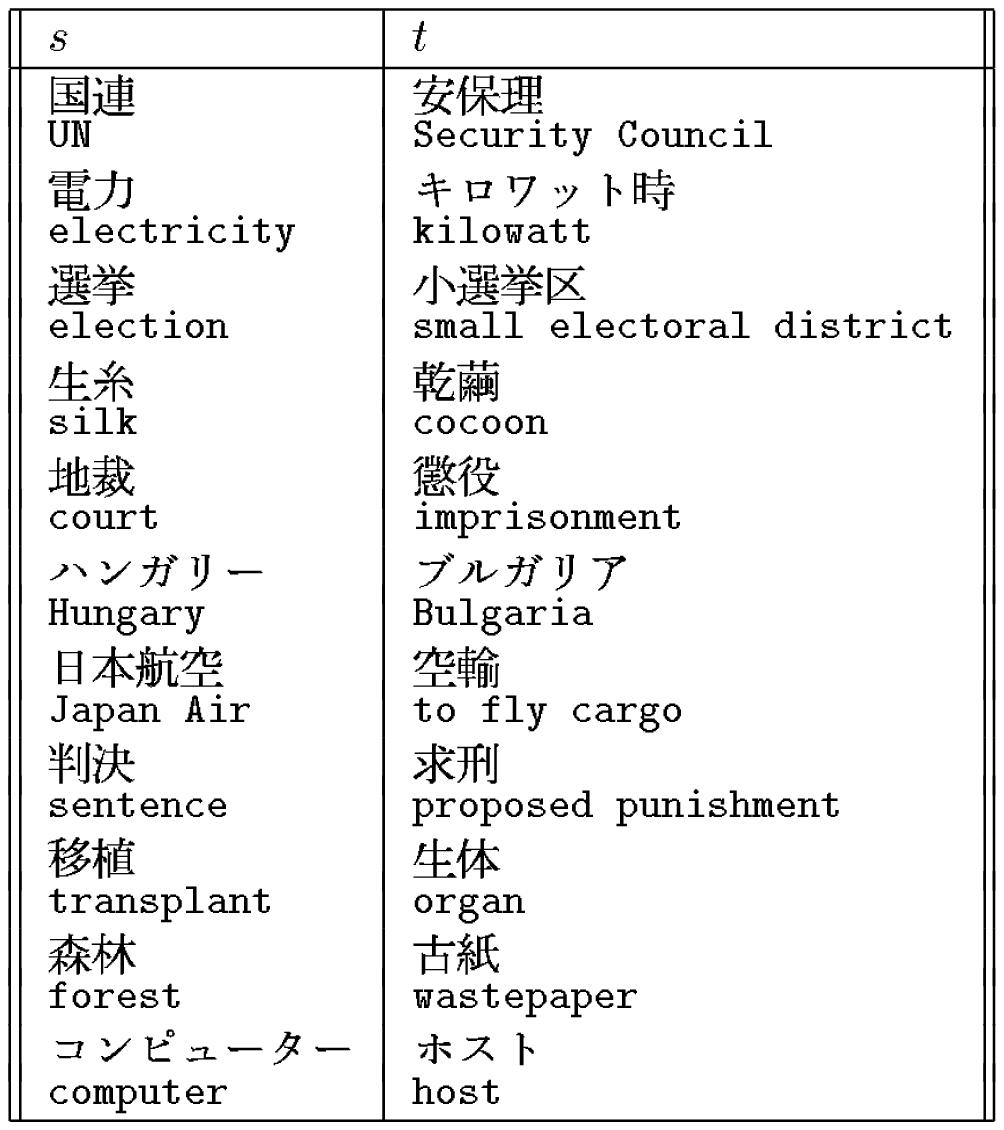,width=3.2in}}
\mbox{}
\vskip-0in
\caption{A sample of triggers extracted from
the 33~million word Nikkei corpus.}
\end{figure}

In earlier work, for example \cite{Rosenfeld:96}, the distance between
the words of a trigger pair $(s,t)$ plays no role in the model, meaning
that the ``boost'' in probability which $t$ receives following its
trigger $s$ is independent of how long ago $s$ occurred, so long as $s$
appeared {\em somewhere} in the {\it history\/} $H$, a fixed-length
window of words preceding $t$.  
It is reasonable to expect, however, that the relevance of a word $s$ 
to the identity of the next word should decay as $s$
falls further and further back into the context. Indeed, there are tables in
\cite{Rosenfeld:96} which suggest that this is so, and
distance-dependent ``memory weights'' are proposed in \cite{Ney:94}.
We decided to investigate the effect of
distance in more detail, and were surprised by what we~found. 

\begin{figure*}[ht]
\begin{center}
\hskip-5pt 
\centerline{
\psfig{file=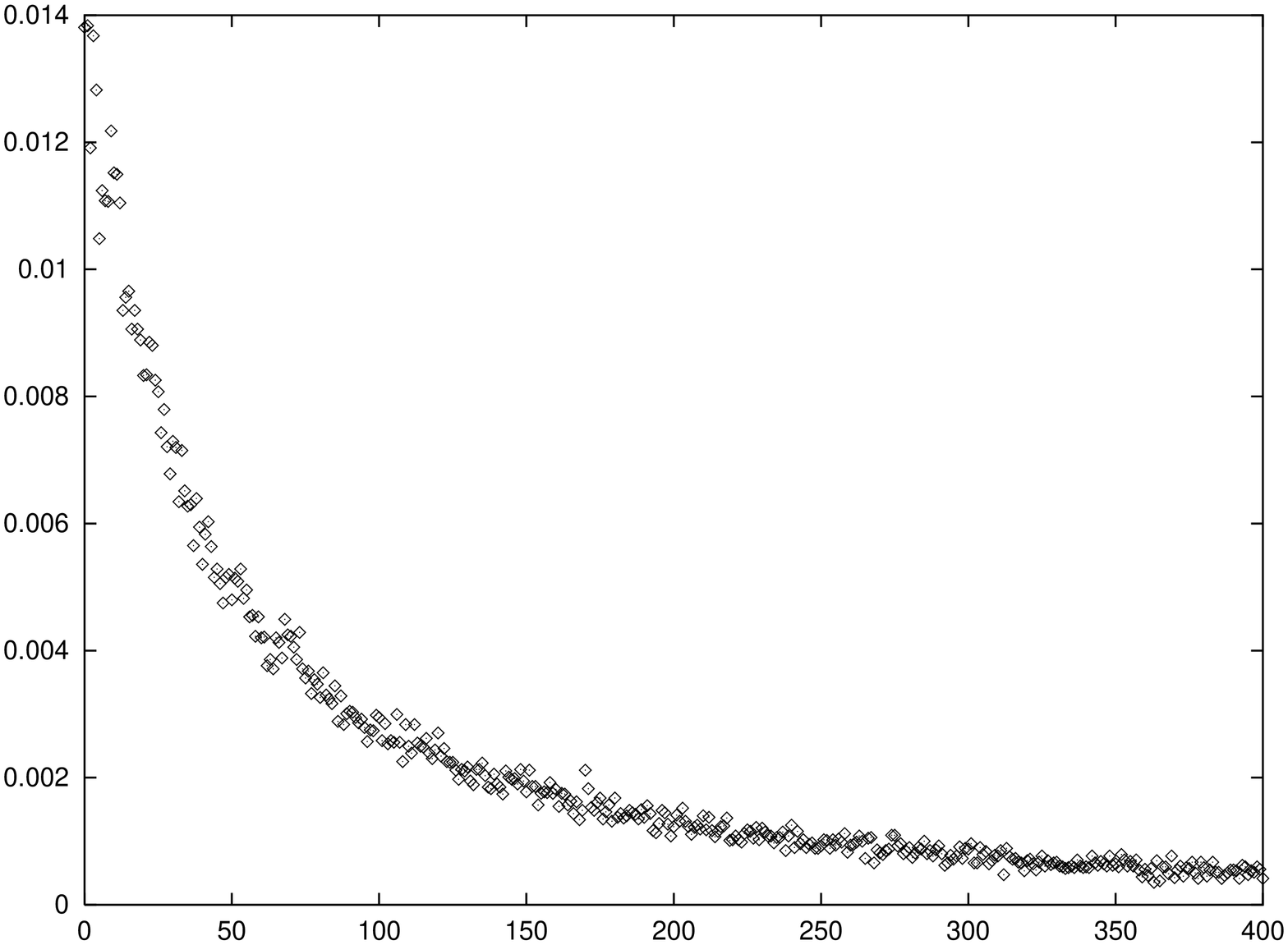,scale=30,angle=-90}
\psfig{file=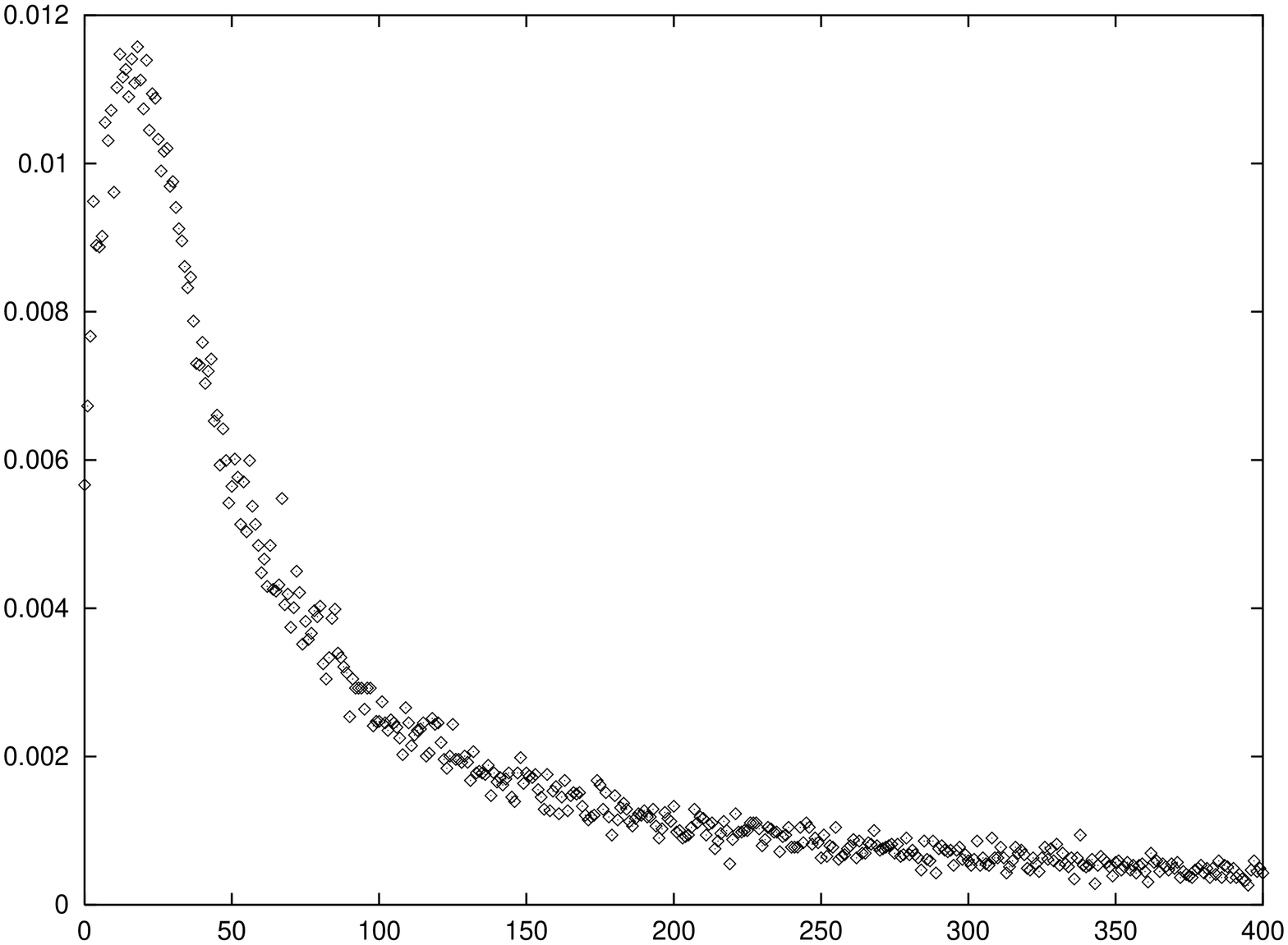,scale=30,angle=-90}}
\caption{The observed distance distributions---collected from
five million words of the Wall Street Journal corpus---for one of the
non-self trigger groups (left) and one of the self trigger groups (right).
For a given distance $0\leq k \leq 400$ on the $x$-axis, the value on 
the $y$-axis is the empirical probability that two trigger words
within the group are separated by exactly $k+2$ words,
conditional on the event that they co-occur within a $400$ word window.
(We exclude separation of one or two words because of our
use of distance models to improve upon trigrams.)}
\label{fig:wsj-full-plots}
\end{center}
\end{figure*}

The set of $41,263$ trigger pairs was partitioned into 20 groups of {\it
non-self} triggers $(s,t)$, $s\neq t$, such as ({\tt Soviet, Kremlin's}), 
and 20 groups of self triggers $(s,s)$, such as ({\tt
business, business}). Figure~\ref{fig:wsj-full-plots} displays the 
empirical probability that
a word $t$ appears for the first time $k$ words after the appearance of
its mate $s$ in a trigger pair $(s,t)$, for two representative groups.

The curves are striking in both their similarities and their
differences.  Both curves seem to have more or less flattened out by
$\W=400$, which allows us to make the approximating assumption (of great
practical importance) that word-triggering effects may be neglected after
several hundred words. The most prominent distinction
between the two curves is the peak near $k=25$ in the self trigger plots;
the non-self trigger plots suggest a monotonic decay.  The shape of the self
trigger curve, in particular the rise between $k=1$ and $k\approx 25$,
reflects the stylistic and syntactic injunctions against repeating a
word too soon. This effect, which we term the {\it lexical exclusion
principle\/}, does not appear for non-self triggers.  In general, the 
lexical exclusion principle seems to be more in effect for uncommon words, 
and thus the peak for such words is shifted further to the right. 

\begin{figure*}[ht]
\begin{tabular}{cccc}
\hskip-10pt 
\psfig{file=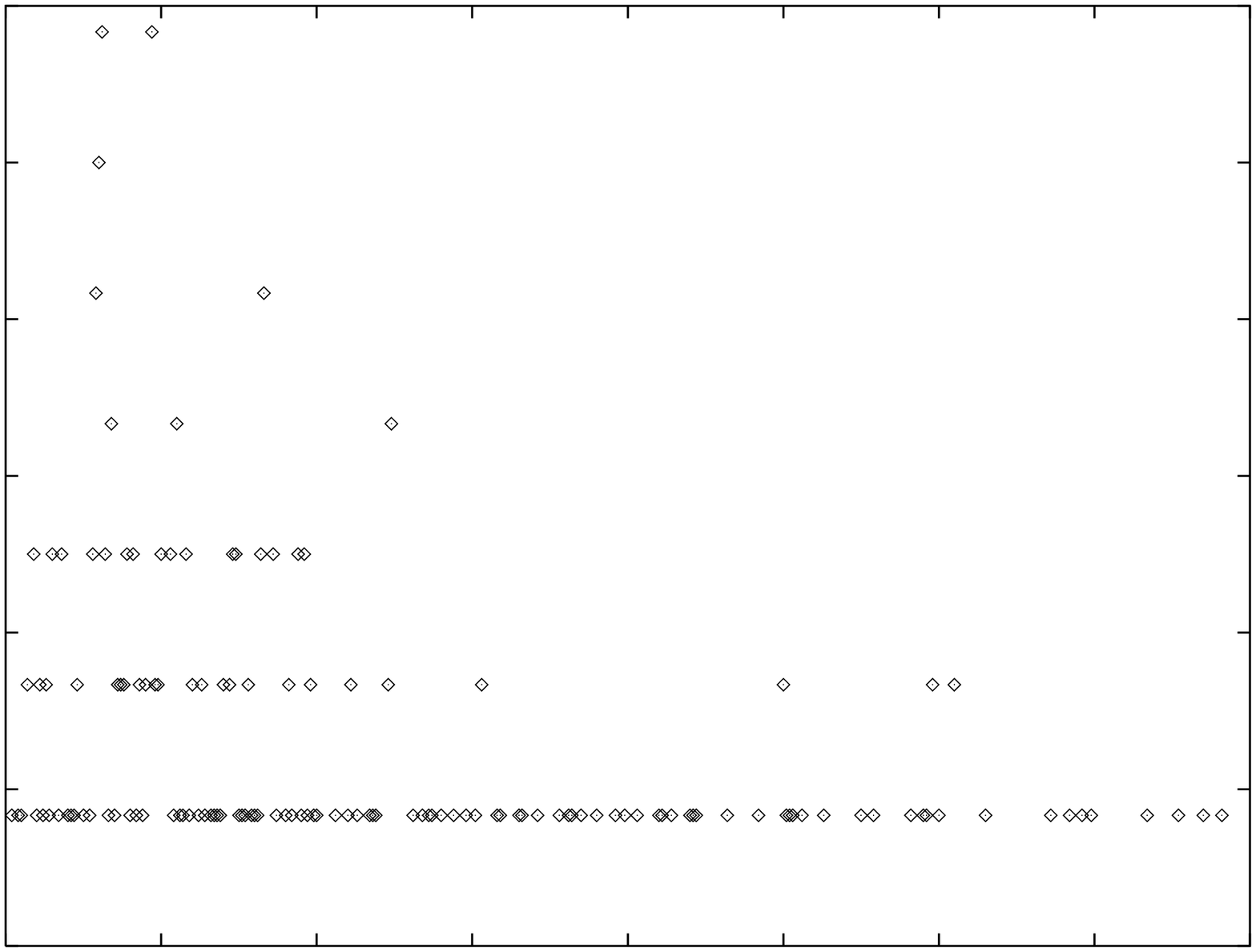,scale=15,angle=-90} &
\psfig{file=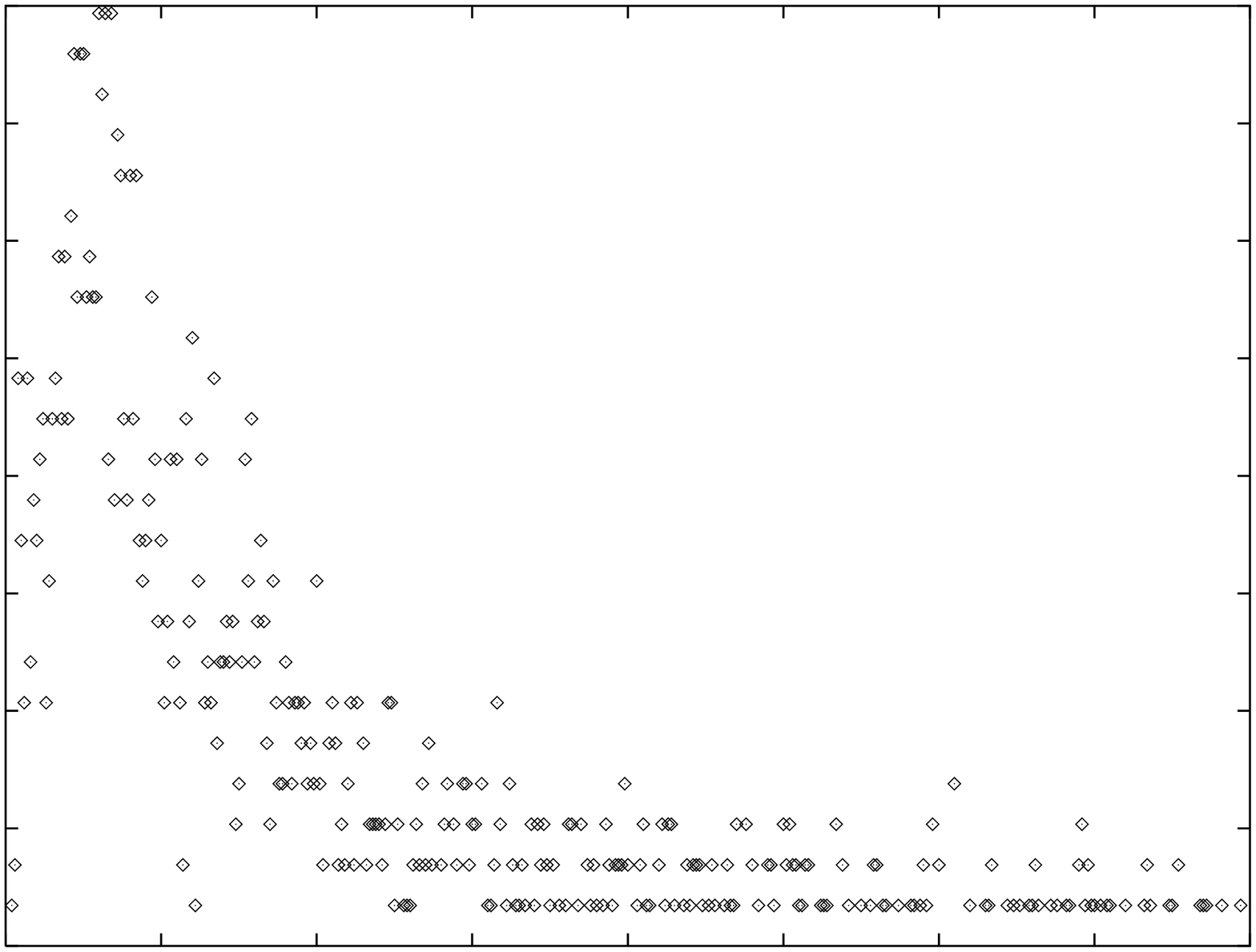,scale=15,angle=-90} &
\psfig{file=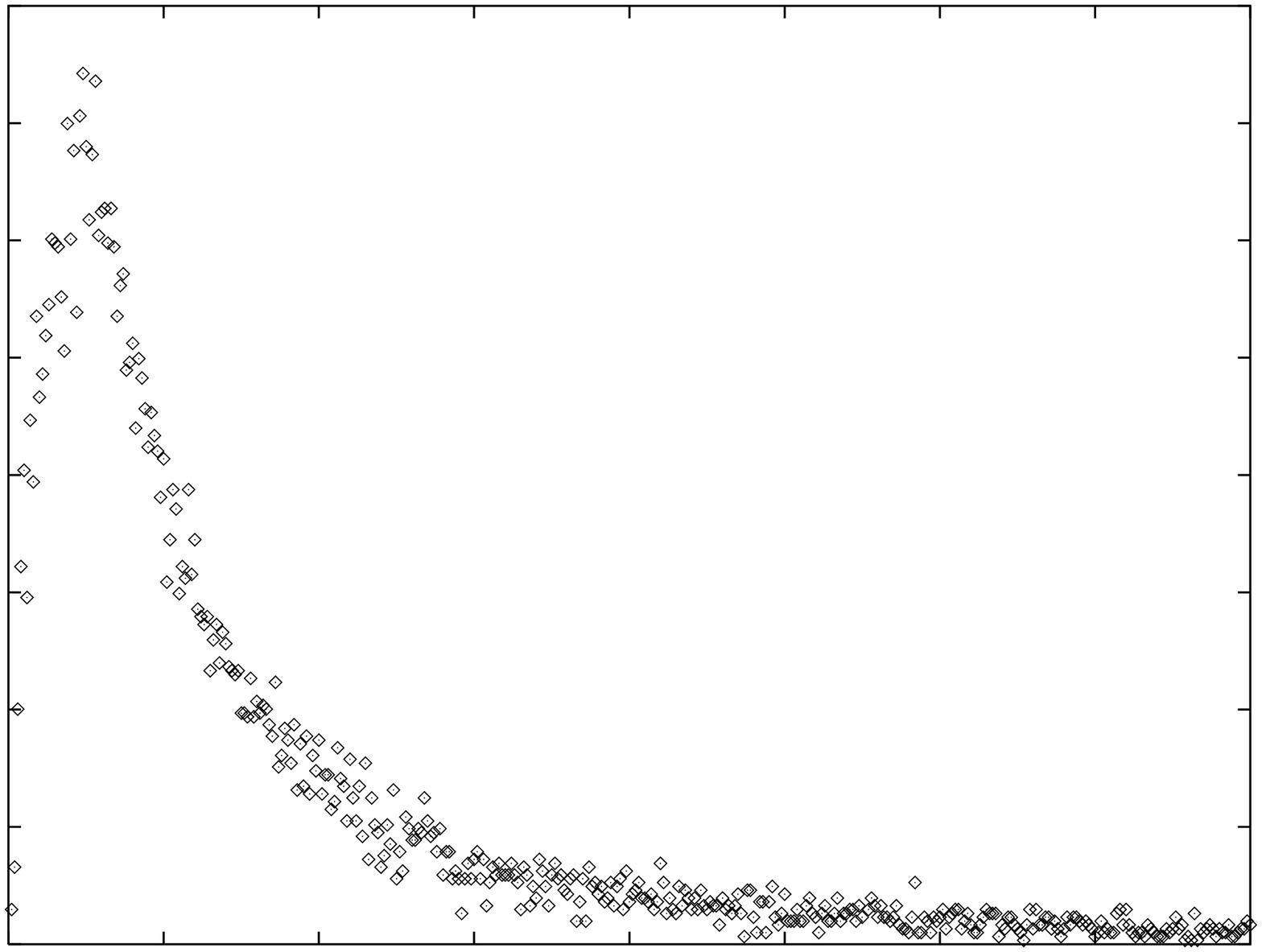,scale=15,angle=-90} &
\psfig{file=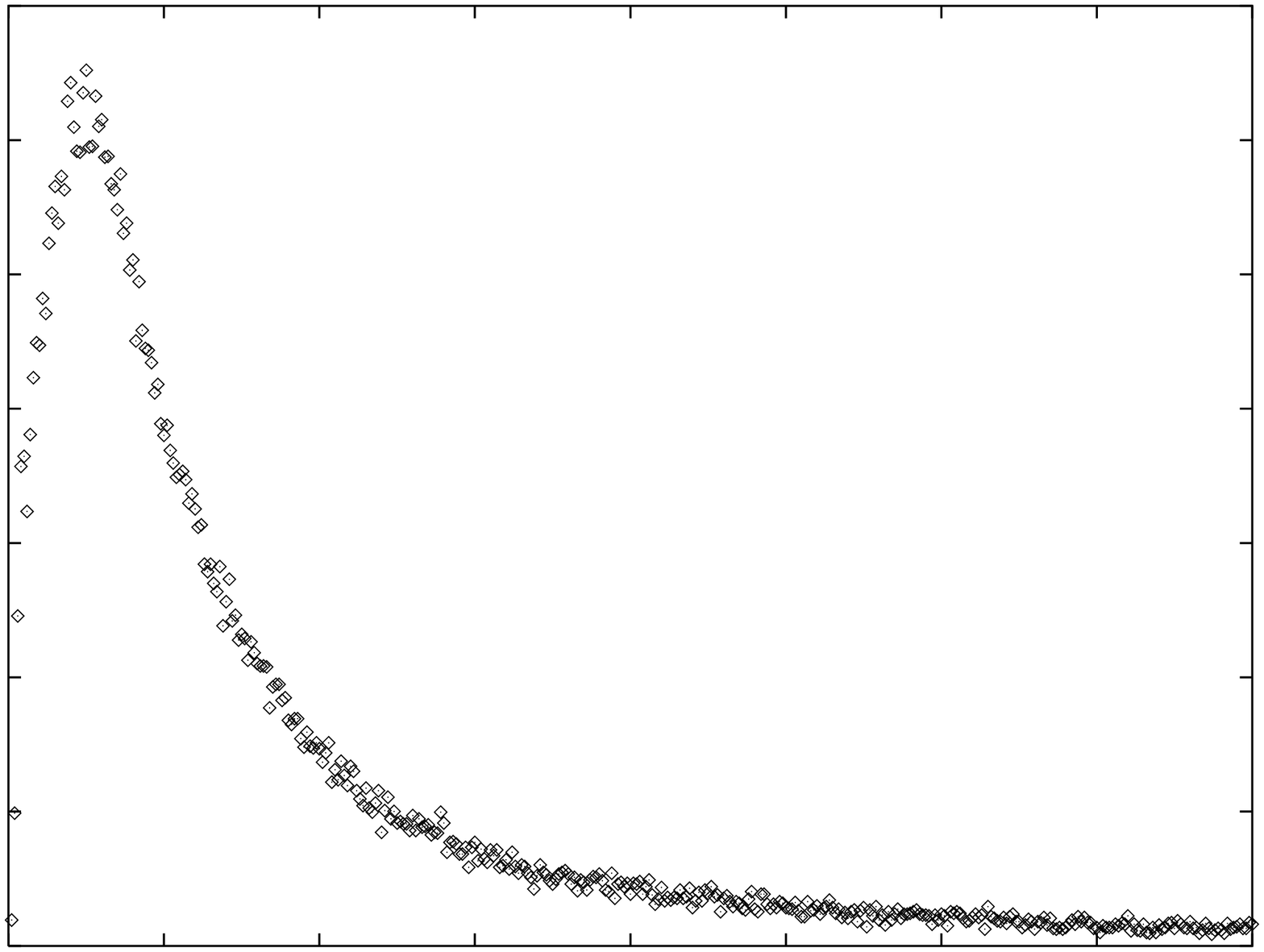,scale=15,angle=-90} 
\end{tabular}
\caption{The law of large numbers emerging for distance
distributions.  Each plot shows the empirical distance curve
for a collection of self triggers, each of which appears fewer
than 100 times in the entire 38 million word Wall Street Journal 
corpus. The plots include statistics for 10, 50, 500, and all 2779 of the
self triggers which occurred no more than 100 times each.}
\label{fig:large-numbers}
\end{figure*}

While the details of the curves vary depending on the particular
triggers, this behavior appears to be universal.  For triggers that
appear too few times in the data for this behavior to exhibit itself,
the curves emerge when the counts are pooled with those from a
collection of other rare words.  An example of this law of large numbers
is shown in Figure~\ref{fig:large-numbers}.

\begin{figure*}[ht]
\begin{tabular}{ccc}
\hskip-5pt 
\psfig{file=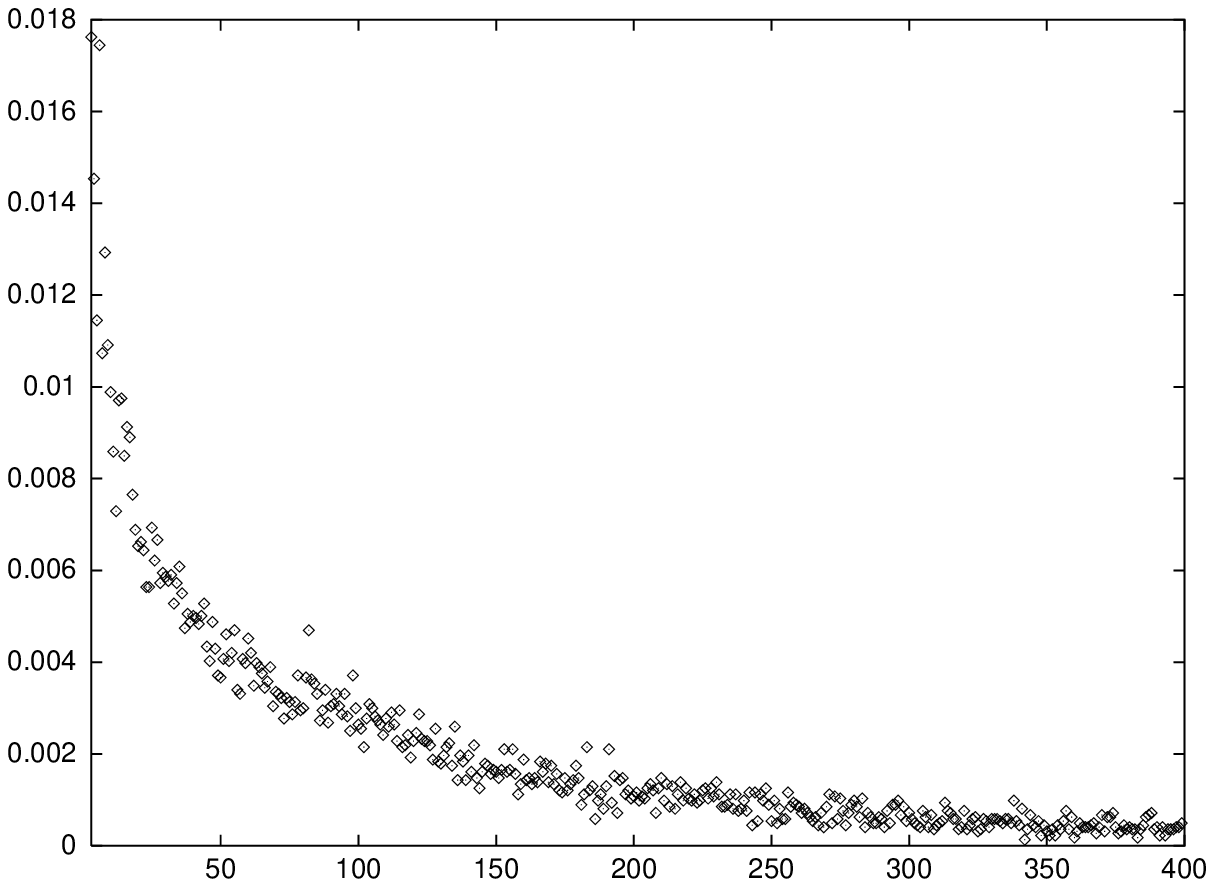,scale=40} &
\psfig{file=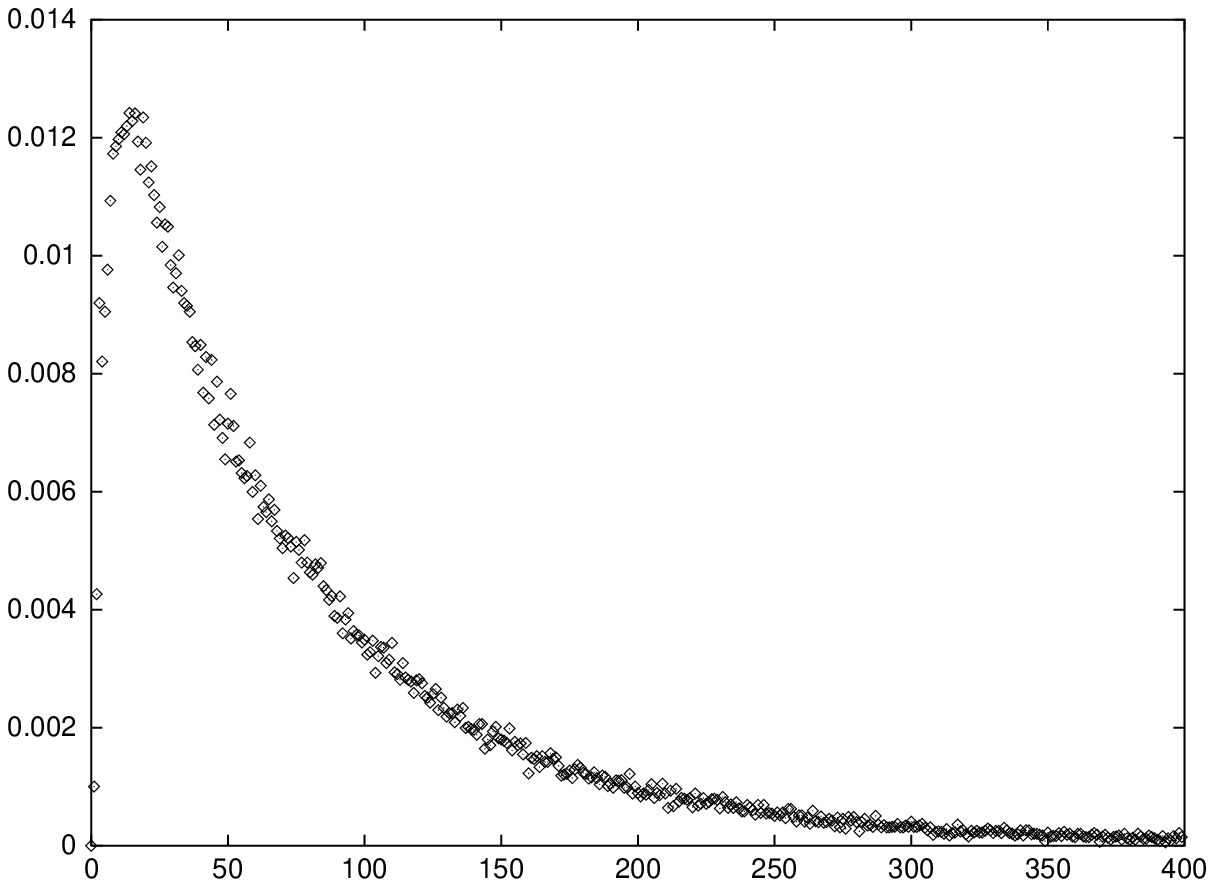,scale=40} &
\psfig{file=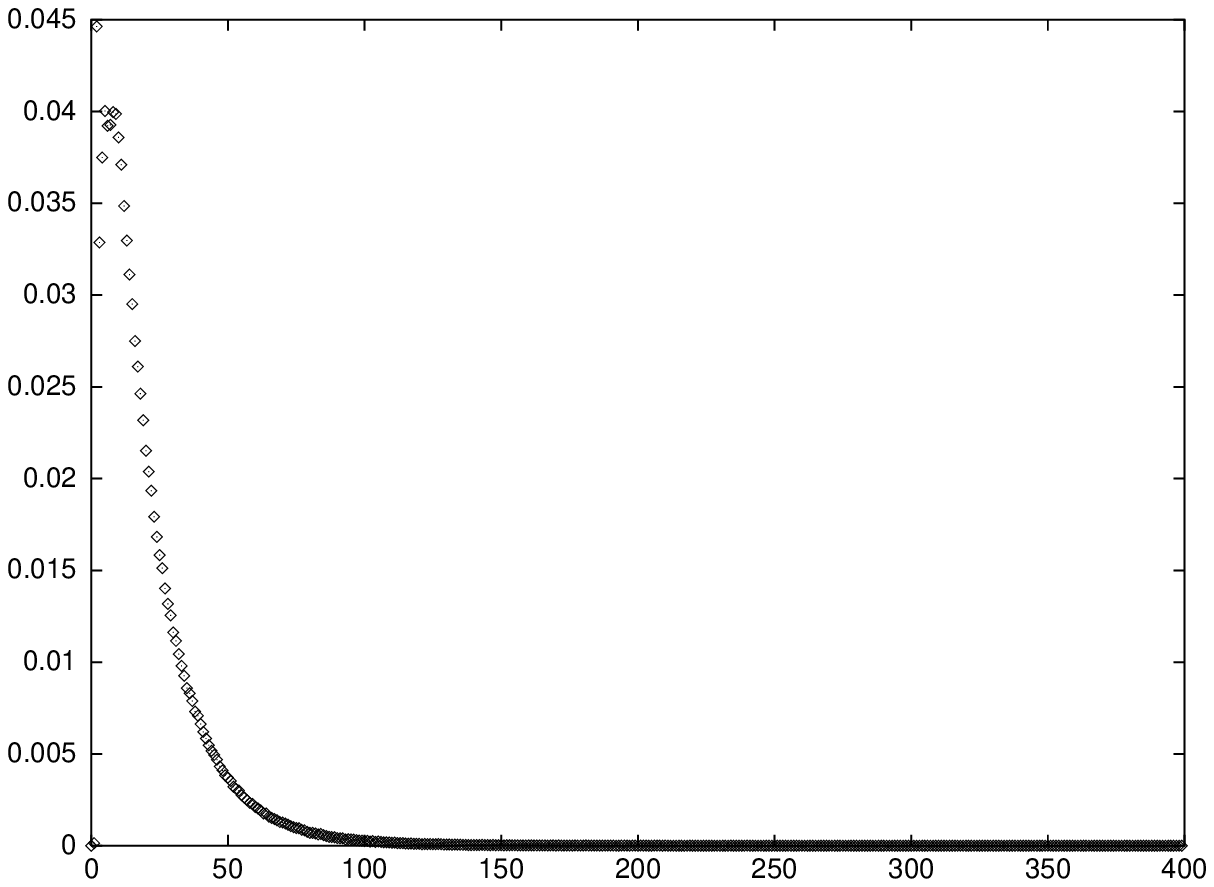,scale=40} \\
&&\cr\medskip
\hskip-5pt 
\psfig{file=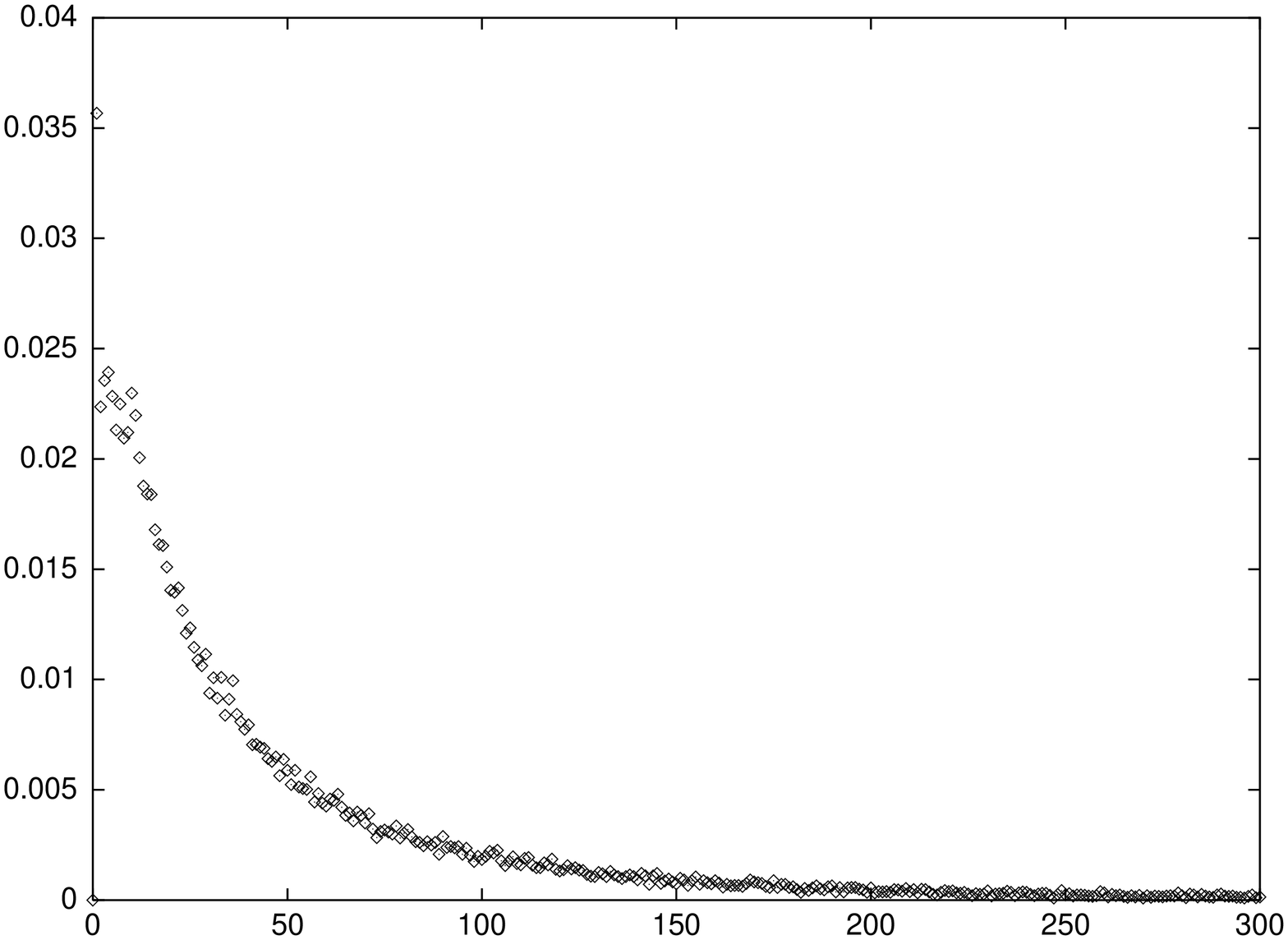,scale=20,angle=-90} &
\psfig{file=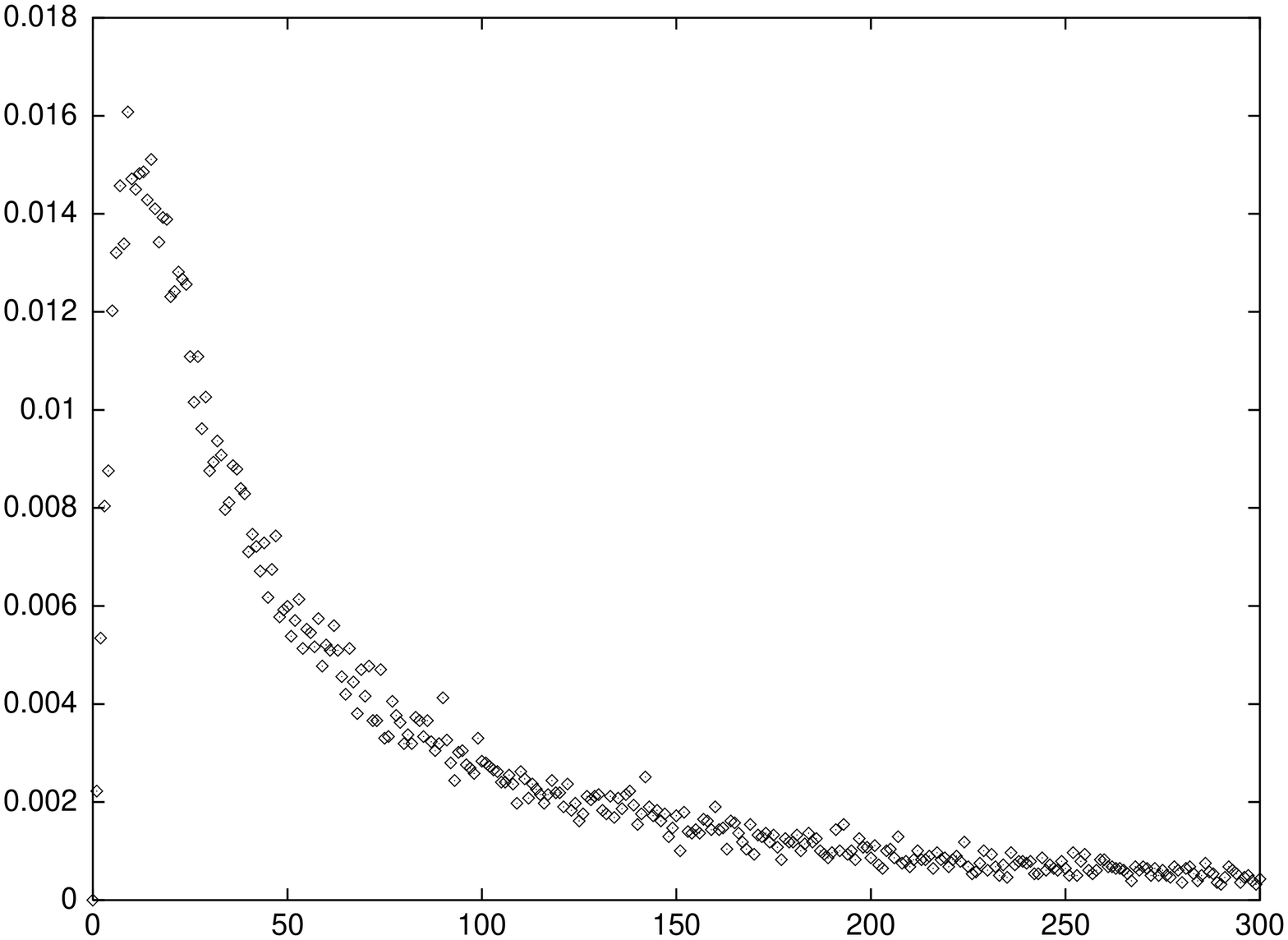,scale=20,angle=-90} &
\psfig{file=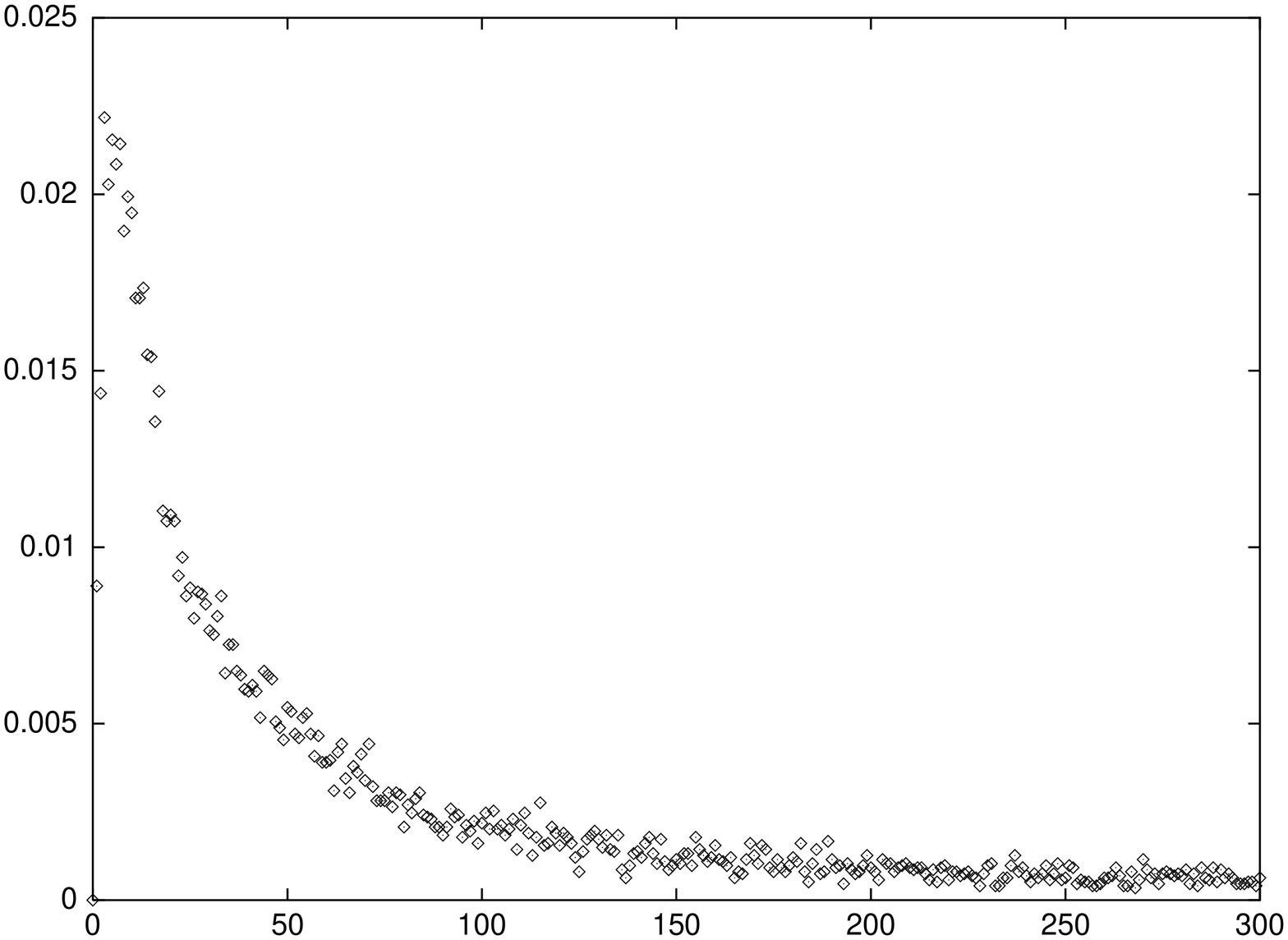,scale=20,angle=-90} 
\end{tabular}
\caption{Empirical distance distributions of triggers
in the Japanese Nikkei corpus, and the Switchboard corpus of conversational speech.
Upper row: All non-self (left) and self triggers (middle)
appearing fewer than 100 times in the Nikkei corpus, and the
curve for the possessive particle {\it no\/} (right).
Bottom row: self trigger {\tt UH} (left), {\tt YOU-KNOW} (middle),
 and all self triggers appearing fewer than 100 times in the entire 
Switchboard corpus (right).}
\label{fig:swb-plots}
\end{figure*}

These empirical phenomena are not restricted to the Wall Street Journal
corpus. In fact, we have observed similar behavior in conversational
speech and Japanese text.  
The corresponding data for self triggers in the Switchboard
data \cite{Godfrey:92a}, for instance, exhibits the same bump in $p(k)$
for small $k$, though the peak is closer to zero.  The lexical exclusion
principle, then, seems to be less applicable when two people are
conversing, perhaps because the stylistic concerns of written
communication are not as important in conversation.
Several examples from the 
Switchboard and Nikkei corpora are shown in Figure~\ref{fig:swb-plots}.

\section{Exponential Models of Distance}

The empirical data presented in the previous section exhibits three salient
characteristics. First is the decay of the probability of a word $t$ as the
distance $k$ from the most recent occurrence of its mate $s$ increases. 
The most important (continuous-time) distribution with this property is the
single-parameter exponential family
\begin{displaymath}
p_\mu(x) = \mu \, e^{-\mu x} \,.
\end{displaymath}
(We'll begin by showing the continuous analogues of the discrete formulas we actually use,
since they are simpler in appearance.)  This family is uniquely characterized
by the {\it memoryless property\/} that the probability of waiting an additional
length of time $\Delta t$ is independent of the time elapsed so far, and
the distribution $p_\mu$ has mean $1/\mu$ and variance $1/\mu^2$.  This
distribution is a good candidate for modeling non-self triggers.

\begin{figure}[ht]
\vskip20pt
\centerline{\psfig{file=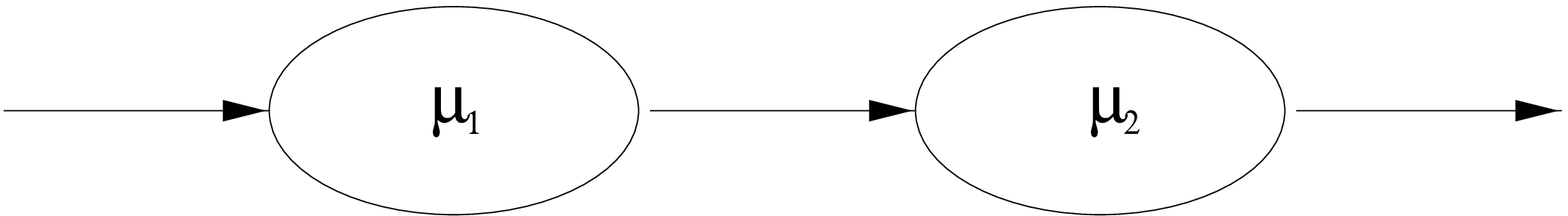,scale=30}}
\caption{A two-stage queue}
\label{fig:twostage}
\end{figure}

The second characteristic is the bump between $0$ and $25$ words
for self triggers.  This behavior appears when two exponential distributions
are arranged in serial, and such distributions are an important tool in
the ``method of stages'' in queueing theory \cite{Kleinrock:75a}. 
The time it takes to travel through two service facilities arranged 
in serial, where the first provides exponential service with rate
$\mu_1$ and the second provides exponential service with rate
$\mu_2$, is simply the convolution of the two exponentials:
\begin{eqnarray*}
p_{\mu_1, \mu_2} (x) & = & \mu_1 \mu_2 \int_{0}^{x} e^{-\mu_1 t} e^{-\mu_2
(x-t)} \, dt \\
& = & {\mu_1 \mu_2 \over \mu_2 - \mu_1} \left( e^{-\mu_1 x} -
e^{-\mu_2 x}\right)\quad \mu_1 \neq \mu_2\,.
\end{eqnarray*}

\begin{figure}[hb]
\centerline{\psfig{file=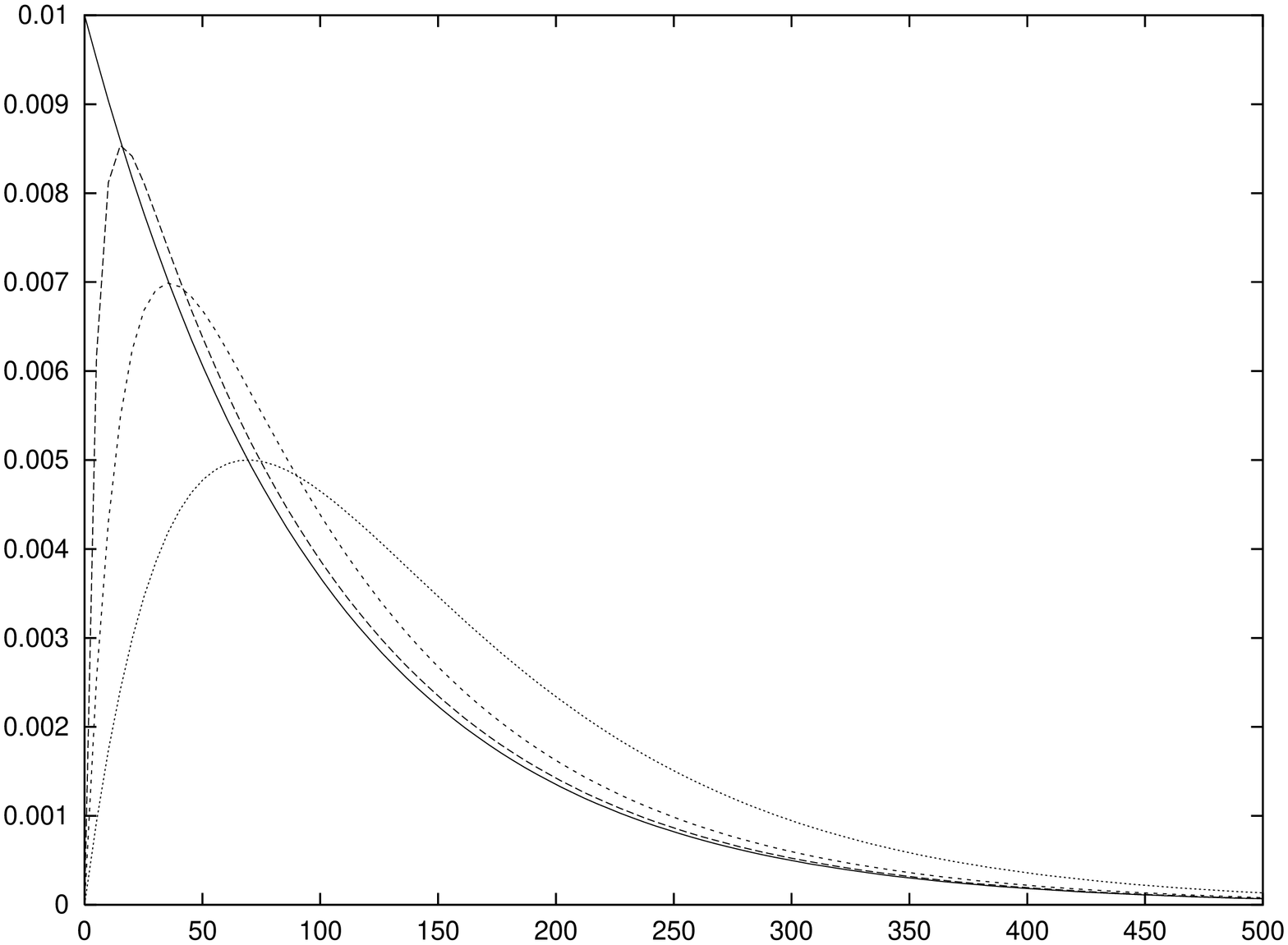,scale=30,angle=-90}}
\caption{A sequence of two-stage exponential models $p_{\mu_1,\mu_2}(x)$ with
$\mu_1 = 0.01,\, 0.02,\, 0.06,\, 0.2,\, \infty$ and $\mu_2 = 0.01$.}
\label{fig:2-stage-sequence}
\end{figure}

 The mean
and variance of the two-stage exponential $p_{\mu_1, \mu_2}$ are
$1/\mu_1 + 1/\mu_2$ and  $1/\mu_1^2 + 1/\mu_2^2$ respectively.
As $\mu_1$ (or, by
symmetry, $\mu_2$) gets large, the peak shifts
towards zero and the distribution approaches the single-parameter exponential
$p_{\mu_2}$ (by symmetry, $p_{\mu_1}$).   A sequence of two-stage models
is shown in Figure~\ref{fig:2-stage-sequence}.

The two-stage exponential is a good candidate for distance modeling
because of its mathematical properties, but it is also well-motivated
for linguistic reasons.  The first queue in the two-stage model
represents the stylistic and syntactic constraints that prevent a word
from being repeated too soon.  After this waiting period, the
distribution falls off exponentially, with the memoryless property. For
non-self triggers, the first queue has a waiting time of zero,
corresponding to the absence of linguistic constraints against using $t$
soon after $s$ when the words $s$ and $t$ are different.  Thus, we are
directly modeling the ``lexical exclusion'' effect and long-distance
decay that have been observed empirically.

The third artifact of the empirical data is the tendency of the
curves to approach a constant, positive value for large distances.
While the exponential distribution quickly approaches zero, the
empirical data settles down to a nonzero steady-state value.

Together these three features suggest modeling distance with a 
three-parameter family of distributions: 
\begin{displaymath}
p_{\mu_1, \mu_2, c}(x) = \gamma^{-1} (\p_{\mu_1,\mu_2}(x) + c)
\end{displaymath}
where $c>0$ and $\gamma$ is a normalizing constant.  Rather than a
continuous-time exponential, we use the discrete-time analogue
\begin{displaymath}
p_{\mu}(k) = (1-e^{-\mu})\, e^{-\mu k}\,.
\label{discrete-exp}
\end{displaymath}
In this case, the two-stage model becomes the discrete-time convolution 
\begin{displaymath}
\p_{\mu_1,\mu_2}(k) =
   \sum_{t=0}^{k} \, p_{\mu_1}(t) \, p_{\mu_2}(k-t) \,.
\end{displaymath}

\noindent{\bf Remark.} \enspace It should be pointed out that there is another
parametric family that is an excellent candidate for distance models, based on the
first two features noted above.  This is the {\it Gamma distribution\/}
\begin{displaymath}
p_{\alpha,\beta}(x) = {\beta^{\alpha} \over \Gamma(\alpha)} \, x^{\alpha-1} e^{-\beta x}\,.
\end{displaymath}
This distribution has mean $\alpha/\beta$ and variance $\alpha/\beta^2$
and thus can afford greater flexibility in fitting the empirical data.
For Bayesian analysis, this distribution is appropriate as the
conjugate prior for the exponential parameter $\mu$ \cite{Gelman:95a}.
Using this family, however, sacrifices the linguistic interpretation of the
two-stage model.

\section{Estimating the Parameters}

In this section we present a solution to the problem of estimating the 
parameters of the distance models introduced in the previous section.
We use the maximum likelihood criterion to fit the curves.  
Thus, if $\theta\in\Theta$ represents the
parameters of our model, and $\pt(k)$ is the empirical probability that
two triggers appear a distance of $k$ words apart, then we seek to
maximize the log-likelihood
\begin{displaymath}
\LL(\theta) = \sum_{k\geq 0} \pt(k) \log p_\theta(k)\,.
\end{displaymath}
First suppose that $\{p_\theta\}_{\theta\in\Theta}$ is the family of
continuous one-stage exponential models $p_\mu(k) = \mu e^{-\mu k}$.
In this case the maximum likelihood problem is straightforward:  the mean
is the sufficient statistic for this exponential family, and its
maximum likelihood estimate is determined by 
\begin{displaymath}
\mu = {1 \over \sum_{k\geq 0} k\, \pt(k)} = {1\over \pmean{\pt}{k}}\,.
\end{displaymath}
In the case where we instead use the discrete model 
$p_\mu(k) = (1-e^{-\mu}) \, e^{-\mu k}$, a little algebra shows that the
maximum likelihood estimate is then
\begin{displaymath}
\mu = \log\left(1 + {1\over\pmean{\pt}{k}}\right)\,.
\end{displaymath}

Now suppose that our parametric family $\{p_\theta\}_{\theta\in\Theta}$
is the collection of two-stage exponential models;  the log-likelihood in
this case becomes
\begin{displaymath}
\label{sum-in-log}
\LL(\mu_1, \mu_2) = \sum_{k\geq 0} \pt(k) 
  \log \left( \sum_{j=0}^{k}  p_{\mu_1}(j) \, p_{\mu_2}(k-j) \right) .
\end{displaymath}
Here it is not obvious how to proceed to obtain the maximum likelihood
estimates.  The difficulty is that there is a sum inside the logarithm,
and direct differentiation results in coupled equations for $\mu_1$ and
$\mu_2$.  Our solution to this problem is to view the convolving index
$j$ as a {\it hidden variable\/} and apply the EM algorithm
\cite{Dempster:77a}.  Recall that the interpretation of $j$ is the
time used to pass through the first queue; that is, the number of
words used to satisfy the linguistic constraints of lexical exclusion.
This value is hidden given only the total time $k$ required to pass
through both queues.

Applying the standard EM argument, 
the difference in log-likelihood for two
parameter pairs $(\mu_1^\prime, \mu_2^\prime)$ and 
$(\mu_1, \mu_2)$ can be bounded from below as
\begin{eqnarray*}
\lefteqn{\LL(\mu^\prime) - \LL(\mu)  = \sum_{k\geq 0} \pt(k) \log\left({ \ptsp(k) \over \pts(k)} \right)} \\
& \geq & \sum_{k\geq 0} \pt (k) \sum_{j=0}^{k} \pts(j\given k) 
\log \left({\ptsp(k,j) \over \pts(k,j)}\right) \\
& \equiv & \A(\mu^\prime, \mu)
\end{eqnarray*}
where 
\begin{displaymath}
p_{\mu_1,\mu_2}(k,j) = p_{\mu_1}(j) \, p_{\mu_2}(k-j) 
\end{displaymath}
and 
\begin{displaymath}
p_{\mu_1,\mu_2}(j \given k) = {p_{\mu_1, \mu_2}(k,j) \over p_{\mu_1,\mu_2}(k)} \,.
\end{displaymath}
Thus, the auxiliary function $\A$ can be written as 
\begin{eqnarray*}
\A(\mu^\prime, \mu) & = & \log\left(1-e^{-\mu_1^\prime}\right) 
	 + \log\left(1-e^{-\mu_2^\prime}\right) \\
	& & - \; \mu_1^\prime \sum_{k\geq 0} \pt(k) \sum_{j=0}^{k} j\, \pts(j\given k) \\
	& & - \mu_2^\prime \sum_{k\geq 0} \pt(k) \sum_{j=0}^{k} (k-j) \, \pts(j\given k) \nonumber\\
	& &+ \; \hbox{\it constant\/}(\mu) \nonumber\,.
\end{eqnarray*}
Differentiating $\A(\mu^\prime, \mu)$ 
with respect to $\mu_i^\prime$, we get the EM updates
$$
\mu_1^\prime = \log\left( 1 + {1 \over  \sum_{k\geq 0} \pt(k)
 	    \sum_{j=0}^{k} j\, \pts(j\given k) } \right) $$
$$ 
\mu_2^\prime = \log\left( 1 + {1 \over \sum_{k\geq 0} \pt(k)
 	    \sum_{j=0}^{k} (k-j)\, \pts(j\given k) } \right) $$

\noindent{\bf Remark.} \enspace It appears that the above updates
require $O(\W^2)$ operations if
a window of $\W$ words is maintained in the history.  However, using
formulas for the geometric series, such as
$\sum_{k=0}^{\infty} k x^k = {x / (1-x)^2}$,
we can write the expectation $\sum_{j=0}^{k} j\, \pts(j\given k)$ in
closed form.  Thus, the updates can be calculated in linear time.

Finally, suppose that our parametric family $\{p_\theta\}_{\theta\in\Theta}$
is the three-parameter collection of two-stage exponential models together
with an additive constant:
\begin{displaymath}
p_{\mu_1, \mu_2, c}(k) = \gamma^{-1} (\p_{\mu_1,\mu_2}(k) + c)\,.
\end{displaymath}
Here again, the maximum likelihood problem can be solved by introducing
a hidden variable.  In particular, by setting
$\alpha = \frac{c}{c + N^{-1}}$
we can express this model as a {\it mixture\/} of a two-stage exponential and a uniform
distribution:
\begin{displaymath}
p_{\mu_1, \mu_2, \alpha}(k) = (1-\alpha)\, \p_{\mu_1,\mu_2}(k) + \alpha \left({1\over \W}\right) \,.
\end{displaymath}
Thus, we can again apply the EM algorithm to determine the mixing
parameter $\alpha$.  This is a standard application of the EM algorithm,
and the details are omitted.

\begin{figure*}[ht]
\begin{tabular}{cc}
\hskip-5pt 
\psfig{file=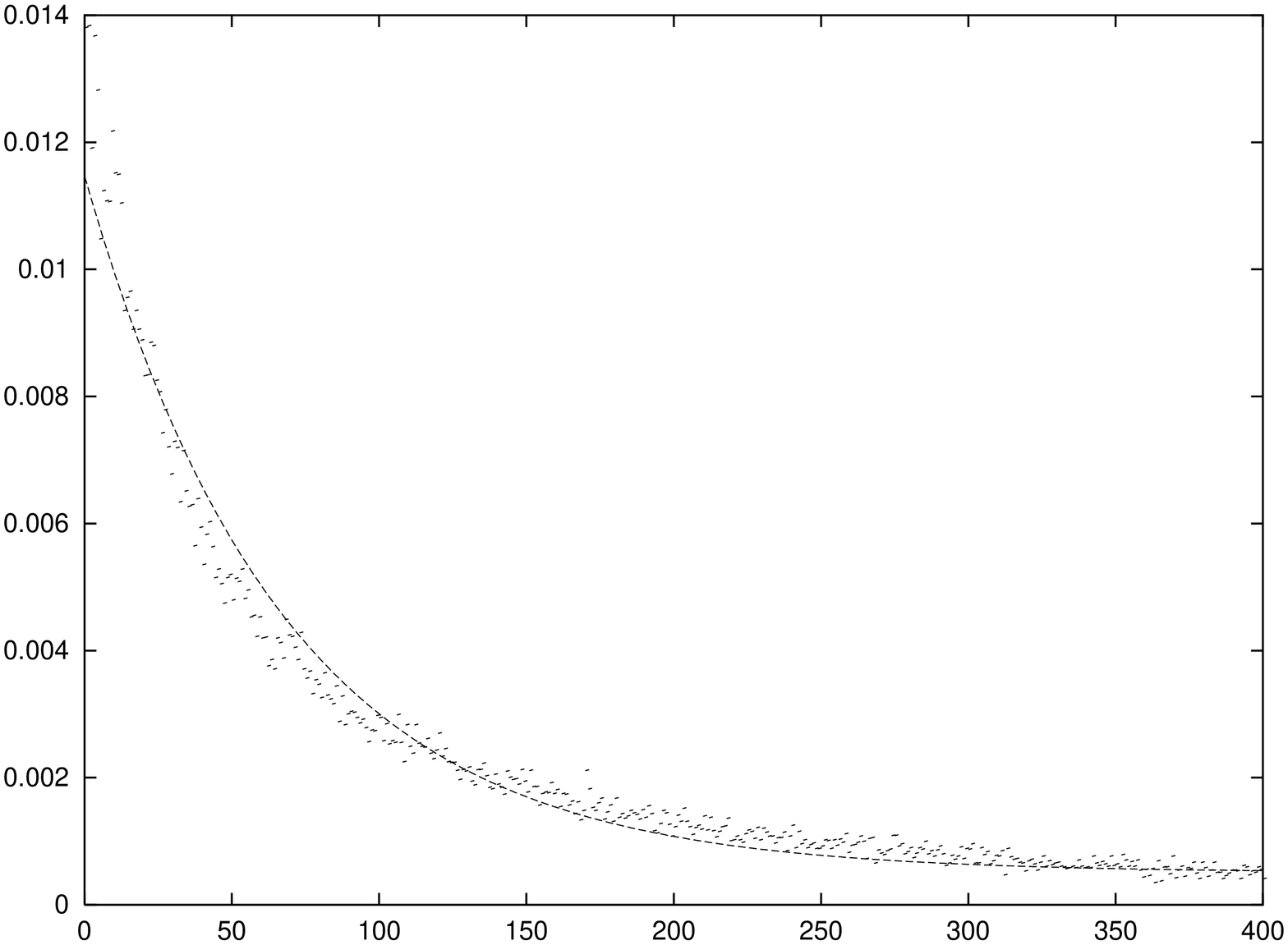,scale=30,angle=-90} &
\psfig{file=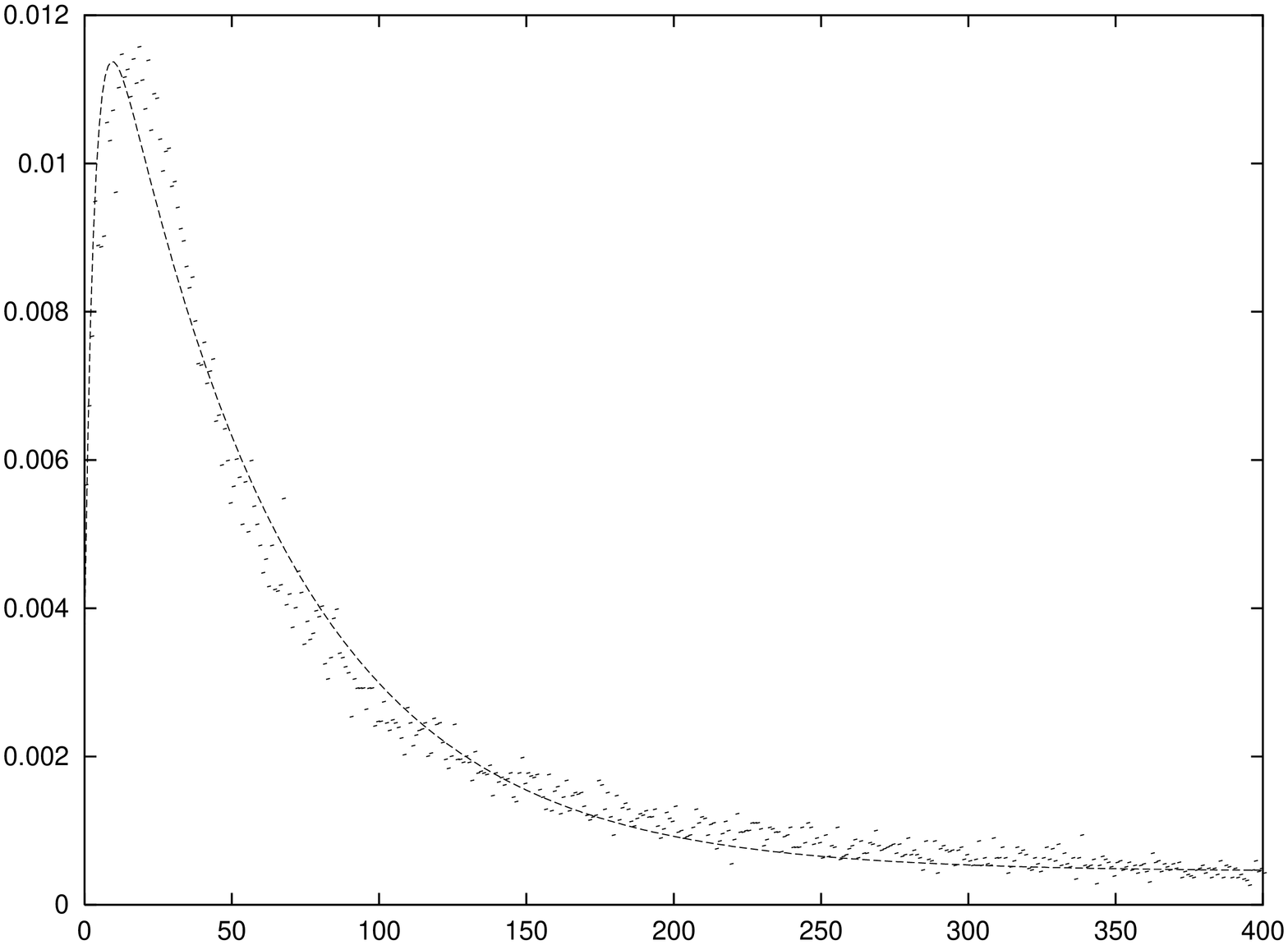,scale=30,angle=-90} \cr
\end{tabular}
\caption{The same empirical distance distributions of Figure~2 fit to the 
three-parameter mixture model $p_{\mu_1, \mu_2, \alpha}$ using the 
EM algorithm. The dashed line is the fitted curve.  For the
non-self trigger plot $\mu_1=7$, $\mu_2=0.0148$, and $\alpha=0.253$.  
For the self trigger plot $\mu_1=0.29$, $\mu_2=0.0168$, and $\alpha=0.224$.}
\label{fig:fits}
\end{figure*}

In summary, we have shown how the EM algorithm can be applied to
determine maximum likelihood estimates of the three-parameter family 
of distance models $\left\{p_{\mu_1, \mu_2, \alpha}\right\}$ 
of distance models.  In Figure~\ref{fig:fits}
we display typical examples of this training algorithm at work.

\section{A Nonstationary Language Model}

To incorporate triggers and distance models into a long-distance
language model, we begin by constructing a standard, static backoff
trigram model \cite{Katz:87a}, which we will denote as 
$q(w_0\given w_{-1}, w_{-2})$.  For the purposes of building a model for
the Wall Street Journal data, this trigram model is quickly
trained on
the entire 38-million word corpus.  We then build a family of
conditional exponential models of the general form
\begin{eqnarray*}
\lefteqn{p(w\given H) =} \\
& & {1\over Z(H)} \exp\left( \sum_{i} \lambda_i f_i(H,w)\right) 
\, q(w\given w_{-1}, w_{-2})
\end{eqnarray*}
where $H = w_{-1}, w_{-2}, \ldots, w_{-\W}$ is the word history,
and $Z(H)$ is the normalization constant
\begin{displaymath}
Z(H) = \sum_{w} \exp\left( \sum_{i} \lambda_i f_i(H, w)\right) \, 
q(w\given w_{-1}, w_{-2})
\end{displaymath}
The functions $f_i$, which depend both on the word history $H$ and the
word being predicted, are called {\it features\/}, and each feature
$f_i$ is assigned a weight $\lambda_i$.  In the models that we built,
feature $f_i$ is an indicator function, testing 
for the occurrence of a trigger pair $(s_i, t_i)$:
\begin{displaymath}
f_{i}(H, w) =  \cases{ 1 & if $s_i\in H$ and $w = t_i$\cr
                              0 & otherwise.} 
\end{displaymath}

The use of the trigram model as a {\it default
distribution\/} \cite{Csiszar:96} in this manner is 
new in language modeling.
(One might also use the term {\it prior\/},
although $q(w\given H)$ is not a prior in the strict Bayesian sense.)
Previous work using maximum entropy methods incorporated trigram constraints as
explicit features \cite{Rosenfeld:96}, using the uniform distribution as
the default model.  There are several advantages to incorporating
trigrams in this way.  The trigram component can be efficiently
constructed over a large volume of data, using standard software or
including the various sophisticated techniques for smoothing that have
been developed.  Furthermore, the normalization $Z(H)$ can be computed
more efficiently when trigrams appear in the default distribution.  For
example, in the case of trigger features, since
\begin{displaymath}
Z(H) = 
1 + \sum_{i} \delta(s_i\in H) (e^{\lambda_i} - 1) q(t_i \given w_{-1}, w_{-2})
\end{displaymath}
the normalization involves only a sum over those words that are
actively triggered.  Finally, assuming robust estimates for the parameters
$\lambda_i$, the resulting model is essentially guaranteed to be
superior to the trigram model. The training algorithm we use for
estimating the parameters is the {\it Improved Iterative Scaling\/}
(IIS) algorithm introduced in \cite{DellaPietra:96a}.

To include distance models in the word predictions, we treat the
distribution on the separation $k$ between $s_i$ and $t_i$ in a
trigger pair $(s_i,t_i)$ as a prior.  Suppose first that our distance
model is a simple one-parameter exponential, $p(k \given s_i\in H,
w=t_i) = \mu_i \, e^{-\mu_i k}$.  Using Bayes' theorem, we can then write
\begin{eqnarray*}
\lefteqn{p(w=t_i \given s_i\in H, s_i = w_{-k})}  \\
&=&  { p(w=t_i \given s_i\in H)\, p(k \given s_i\in H, w=t_i)  \over p(k \given s_i\in H)} \\
&\propto&        e^{\lambda_i -\mu_i k} \, q(t_i \given w_{i-1}, w_{i-2})\,.
\end{eqnarray*}
Thus, the distance dependence is incorporated
as a {\it penalizing feature\/}, the effect of which is to discourage a large
separation between $s_i$ and $t_i$.  A similar interpretation holds
when the two-stage mixture models $p_{\mu_1, \mu_2, \alpha}$ are used to
model distance, but the formulas are more complicated.

In this fashion, we first trained distance models using the algorithm
outlined in Section~4.  We then incorporated the distance models as
penalizing features, whose parameters remained fixed, and proceeded to train
the trigger parameters $\lambda_i$ using the IIS algorithm.  Sample
perplexity results are tabulated in Figure~\ref{results2}.

\begin{figure*}[ht]
\begin{center}
\begin{tabular}{c}
\renewcommand{\arraystretch}{1.4}
\begin{tabular}{|l|c|c|}
\hline 
{\it Experiment} & {\it Perplexity} & {\it Reduction} \\
\hline 
Baseline: trigrams trained on 5M words & 170 & --- \\
\hline 
Trigram prior + 41,263 triggers        & 145 & 14.7\% \\ 
\hline 
Same as above + distance modeling      & 142 & 16.5\% \\
\hline
\hline 
Baseline: trigrams trained on 38M words & 107 & --- \\
\hline 
Trigram prior + 41,263 triggers        & 92 & 14.0\% \\ 
\hline 
Same as above + distance modeling      & 90 & 15.9\% \\
\hline
\end{tabular}
\end{tabular}
\end{center}
\caption{Models constructed using trigram priors. Training
the larger model required about 10 hours on a DEC Alpha workstation.}
\label{results2}
\end{figure*}

One important aspect of these results is that because a smoothed trigram
model is used as a default distribution, we are able to {\it
bucket\/} the trigger features and estimate their parameters on a modest
amount of data.  The resulting calculation takes only
several hours on a standard workstation, in comparison to the
machine-months of computation that previous language models of this
type required.  

The use of distance penalties gives only a small improvement, in terms
of perplexity, over the baseline trigger model.  However, we have
found that the benefits of distance modeling can be sensitive to
configuration of the trigger model.  For example, in the results
reported in Table~\ref{results2}, a trigger is only allowed to 
be active once in any given context.  By instead allowing multiple
occurrences of a trigger $s$ to contribute to the prediction of its
mate $t$, both the perplexity reduction over the baseline trigram
and the relative improvements due to distance modeling are increased.

\section{Conclusions}

We have presented empirical evidence showing that the distribution of
the distance between word pairs that have high mutual information
exhibits a striking behavior that is well modeled by a three-parameter
family of exponential models.  The properties of these co-occurrence
statistics appear to be exhibited universally in both text and
conversational speech.  We presented a training algorithm for this class
of distance models based on a novel application of the EM algorithm.
Using a standard backoff trigram model as a default distribution, we
built a class of exponential language models which use nonstationary features
based on trigger words to allow the model to adapt to the recent
context, and then incorporated the distance models as penalizing
features.  The use of distance modeling results in an improvement
over the baseline trigger model.

\section*{Acknowledgement}
We are grateful to Fujitsu Laboratories, and in particular to Akira
Ushioda, for providing access to the Nikkei corpus within Fujitsu
Laboratories, and assistance in extracting Japanese trigger pairs.

\nocite{Berger:96a}

\bibliography{repulsion}

\end{document}